\documentclass[aps,pre,groupedaddress]{revtex4}
\usepackage{amsfonts}
\usepackage{mathrsfs}
\usepackage{graphicx}
\usepackage{subfigure}
\usepackage{amsmath}
\usepackage{amssymb}
\usepackage{amsbsy}
\usepackage{bm}

\def\bs{\bm{\sigma}}
\def\bx{\bm{\xi}}
\def\bl{\bm{\lambda}}
\def\E{\mathbb{E}}
\def\D{\mathcal{D}}
\def\S{\mathcal{S}}
\def\X{X_\mu^a}
\def\cmui{\chi_{\mu\rightarrow i}}
\def\cmu{\chi_{\mu}}
\def\tcmu{\tilde{\chi}_{\mu}}
\def\Lmu{\Lambda_{\mu}}
\def\Lmui{\Lambda_{\mu\rightarrow i}}
\def\tLmu{\tilde{\Lambda}_{\mu}}

\begin{document}

\title{Variational mean-field theory for training restricted Boltzmann machines with binary synapses}
\author{Haiping Huang}
\email{huanghp7@gmail.sysu.edu.cn}
\affiliation{PMI Lab, School of Physics,
Sun Yat-sen University, Guangzhou 510275, People's Republic of China}
\date{\today}

\begin{abstract}
Unsupervised learning requiring only raw data is not only a fundamental function of the
cerebral cortex, but also a foundation for a next generation of artificial neural networks. However, 
a unified theoretical framework to treat sensory inputs, synapses and neural activity together is still lacking. The computational obstacle originates
from the discrete nature of synapses, and complex interactions among these three essential elements of learning. Here, we propose a variational mean-field theory
in which the distribution of synaptic weights is considered. The unsupervised learning can then be decomposed into two intertwined steps: a maximization step is carried out as a 
gradient ascent of the lower-bound on the data log-likelihood, in which the synaptic weight distribution is determined by updating variational parameters, and an expectation step is carried out as a message passing procedure on an equivalent or dual
neural network whose parameter is specified by the variational parameters of the weight distribution. Therefore, our framework provides insights on how data (or sensory inputs), synapses
and neural activities interact with each other to achieve the goal of extracting statistical regularities in sensory inputs. This variational
framework is verified in restricted Boltzmann machines
with planted synaptic weights and handwritten-digits learning. 
\end{abstract}

 \maketitle

\textit{Introduction.}---
Searching for hidden features in raw sensory inputs and thus a reasonable explanation of the inputs is a core function of natural intelligence~\cite{Neuron-2017,Zador-2019}. Sensory cortical circuits are able to extract useful information from
noisy inputs because of unsupervised synaptic plasticity shaping well-organized synaptic connections~\cite{Marr-1970,Barlow-1989}. From a neural network perspective, this kind of unsupervised
learning was modeled by
a simple two-layer architecture, namely restricted Boltzmann machine (RBM)~\cite{Smolensky-1986,Freund-1994,Hinton-2002}. One layer is used to receive the sensory inputs, thereby being the visible layer; while
the other layer serves as a hidden representation encoding features in the inputs. No lateral connections exist within each layer, which was designed to avoid costly sampling as in a fully-connected network.
Synaptic plasticity thus refers to the learning process where the connection (weight) strengths between these two layers are adjusted to explain the sensory inputs. 

In the machine learning community, this
learning process can be achieved by
the popular truncated Gibbs sampling (also called contrast divergence algorithm~\cite{Hinton-2002}), which was particularly designed for continuous weight values, and thus a gradient ascent of the data log-likelihood is
mathematically guaranteed~\cite{Bengio-2009}. However, to model unsupervised learning with energy-efficient computation using low-precision synapses (weights)~\cite{morphic-2015,Hubara-2017}, the gradient-based method does not apply due to the discreteness of synapses.
An efficient or valid method of training RBM with low-precision weights was thus thought to be out of reach. 
Several recent works studied computational principles of unsupervised learning with low-precision synapses~\cite{Huang-2016b,Huang-2017,Barra-2017,Monasson-2017,Huang-2018,Barra-2018,Dece-2019,Huang-2019,Hou-2019}.
Due to the complexity of analysis, these works focused on either of practical network architectures and the constraint of arbitrarily many data samples, thereby failing to analyze the interaction among inputs, synaptic plasticity and neural activity within a unified framework. 
Therefore, how this interaction shapes the unsupervised
learning process is still unknown, and moreover, because a typical learning procedure in both biological neural networks and artificial ones must involve 
sensory inputs, synaptic plasticity and neural activity,
understanding this interaction becomes a key to unlock the black box of unsupervised learning, which is not only a fundamental function of 
cerebral cortex~\cite{Marr-1970,Barlow-1989}, but also a foundation for a next generation of artificial neural networks~\cite{Zador-2019}.

Here, we propose a variational principle that assumes a variational distribution of weights. We model the learning process
as a computation of the posterior on the weight parameters given sensory inputs. 
The variational principle finds the best approximation of the posterior within a tractable parametric family.
Remarkably, even though in the presence of both
a generic RBM architecture and arbitrarily many sensory data, the learning process improves an approximate lower-bound on the data log-likelihood. Therefore, our variational principle
overcomes the computational obstacle previously thought to be challenging,
opening a new
path towards understanding how sensory data, synapses and neurons interact with each other during unsupervised learning. In particular,
this principle can be used to test hypotheses or predictions drawn from theoretical studies of
random models of unsupervised learning~\cite{Huang-2016b,Huang-2017,Barra-2017,Monasson-2017,Huang-2018,Barra-2018,Dece-2019,Huang-2019,Hou-2019}.

\begin{figure}
\centering
     \includegraphics[bb=28 609 415 748,width=0.5\textwidth]{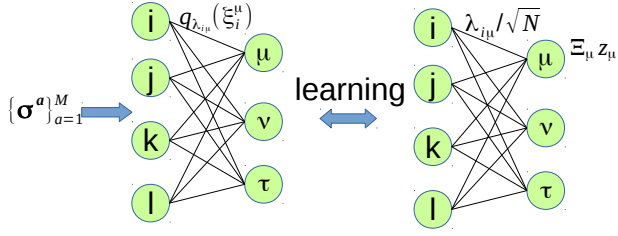}
  \caption{
  (Color online) The schematic illustration of the variational model. $N=4$ in this example (say, $i,j,k$ and $l$). The hidden layer has $P=3$ neurons. Note that $(N,P)$ can be \textit{arbitrarily finite} numbers.
  (Left panel) The RBM architecture receives data of $M$ examples and encodes the hidden features of data into synaptic connections (weights) in an unsupervised way. Here, we take into account the weight uncertainty captured by
  a variational distribution $q_{\lambda_{i\mu}}(\xi_{i}^\mu)$ where $\xi_{i}^\mu$ refers to the connection between sensory neuron $i$ and hidden neuron $\mu$, and $\lambda_{i\mu}$ is the variational 
  parameter.
  (Right panel) The equivalent RBM model where the connection becomes the variational parameter and the additional random field applied to a hidden neuron
  is determined by the variance ($\Xi_\mu^2$) of the weighted-sum input to that hidden neuron (see the main text).
  $z_\mu$ is a standard normally distributed random variable. The equilibrium properties of the equivalent RBM are used to adjust the variational parameter improving the lower-bound on the data log-likelihood.
  }\label{vrbm}
\end{figure}

\textit{Model.}---
In this study, we use the RBM defined above with arbitrarily many hidden neurons (the left panel of Fig.~\ref{vrbm}) to learn hidden features in $M$ input data samples, which are raw unlabeled data specified by $\{\bs^a\}_{a=1}^M$.
Each data sample is
specified by an Ising-like spin configuration $\bs=\{\sigma_i=\pm1\}_{i=1}^{N}$ where $N$ is the size of an input sample (e.g., the number of pixels in an image).
The synaptic weights are characterized by $\bx$, where each component takes a binary value ($\pm1$). The number of hidden neurons is defined by $P$.
The Boltzmann distribution of this RBM is given by $P(\bs)=\frac{1}{Z(\bx)}\prod_{\mu}\cosh(\beta X_\mu)$~\cite{Huang-2015b},
where $\mu$ denotes the hidden neuron index, $X_\mu\equiv\frac{1}{\sqrt{N}}\bx^\mu\cdot\bs$, $\bx^\mu$ is also called the receptive field of the $\mu$-th hidden neuron,
and $Z(\bx)$ is the partition function depending on the joint set of all receptive fields $\bx$.
Note that the hidden neurons' activities ($\pm1$) have been marginalized out. The scaling factor $\frac{1}{\sqrt{N}}$ ensures that the argument is of the order of unity. Supposed that the data samples are 
generated from a RBM where the synaptic connections are randomly generated at first and then quenched.
The inverse-temperature $\beta$ thus tunes the noise level of generated data samples from the planted RBM. Then one standard test of any learning algorithm is to learn the planted
synaptic connection matrix from the supplied data samples.

To further model the learning process, we assume that the data samples are weakly-correlated~\cite{Hinton-2002,Huang-2017}, e.g., sampled with a large interval.
Therefore, we have the following data probability:
\begin{equation}\label{Pbs}
 P(\{\bs^a\}_{a=1}^M|\bx)=\prod_{a=1}^M\frac{1}{Z(\bx)}\prod_\mu\cosh(\beta\X),
\end{equation}
where the superscript $a$ in $\X$ means that $\bs$ in $X_\mu$ is replaced by $\bs^a$. Finally, the learning process is modeled by estimating 
 the posterior probability of the guessed synaptic weights according to the Bayes' rule~\cite{Huang-2017,Huang-2019}:
\begin{equation}\label{Pobs}
\begin{split}
 &P(\bx|\{\bs^{a}\}_{a=1}^{M})=\frac{\prod_{a}P(\bs^{a}|\bx)}{\sum_{\bx}\prod_{a}P(\bs^{a}|\bx)}\\
 &=\frac{1}{\Omega}\exp\left(-M\ln Z(\bx)+\sum_{a,\mu}\ln\cosh\left(\beta\X\right)\right),
 \end{split}
 \end{equation}
where $\Omega$ is the partition function of the posterior, and
a uniform prior for $\bx$ is assumed, i.e., we have no prior knowledge about the planted weights. For simplicity,
we use the same temperature as that used to generate data. 
However, one obstacle to compute the posterior probability is the nested partition function $Z(\bx)\equiv\sum_{\bs}\prod_\mu\cosh(\beta X_\mu)$ which involves an exponential computational complexity.
Except for a few special cases of one or two hidden neurons as studied in previous works~\cite{Huang-2016b,Barra-2017,Huang-2017,Huang-2018,Huang-2019,Hou-2019}, the computation of the posterior is impossible, let alone understanding the learning
process. This is the very \textit{motivation} of this work that proposes a new principled method to tackle this challenge for paving a way towards a scientific understanding of a generic unsupervised learning process.

\textit{A variational principle.}---
Rather than finding an approximate method to evaluate the posterior, we use a variational distribution belonging to the mean-field family~\cite{Saul-96,Jordan-1999,Blei-2017}, defined by $q_{\bl}(\bx)$, and find the best
variational distribution to minimize the Kullback-Leibler (KL) divergence between $q_{\bl}(\bx)$ and $P(\bx|\D)$ where $\D$ denotes the data $\{\bs^a\}$, which is given by
\begin{equation}\label{KL}
\begin{split}
 &{\rm KL}(q_{\bl}(\bx)\| P(\bx|\D))=\E\ln q_{\bl}(\bx)-\E\ln P(\bx|\D)\\
 &=\E\ln q_{\bl}(\bx)-\E\ln P(\bx,\D)+\ln P(\D)\\
 &=-{\rm LB}(q_{\bl})+\ln P(\D),
 \end{split}
 \end{equation}
 where ${\rm KL}(q\|p)\equiv\left<\ln\frac{q}{p}\right>_q$ for two distributions--- $q$ and $p$, $\E$ denotes the expectation under the variational probability $q_{\bl}$ where $\bl$ denotes the corresponding
 variational parameter, and $P(\bx,\D)=P(\bx|\D)P(\D)$. Because of the non-negativity of the KL divergence,
 the lower-bound on the data log-likelihood is given by
 \begin{equation}\label{LB}
 \begin{split}
  {\rm LB}(q_{\bl})&\equiv\E\ln P(\bx,\D)-\E\ln q_{\bl}(\bx)\\
  &=\E\ln P(\D|\bx)-{\rm KL}(q_{\bl}(\bx)\|P(\bx)),
  \end{split}
 \end{equation}
 where the argument of $q_{\bl}$ is omitted when it is clear, and $P(\bx,\D)=P(\D|\bx)P(\bx)$ where $P(\bx)$ can be treated as a prior probability of weights.
The first term serves as the expected log-likelihood of the data, while the second term is a regularization term. The first term encourages the variational distribution to explain 
the observed data (maximizing the expected log-likelihood of the data), while the second term encourages the approximate posterior to match the prior. Therefore,
the variational objective in Eq.~(\ref{LB}) takes into account the balance between likelihood and prior~\cite{Blei-2017}. 
The learning process is now interpreted as finding the variational parameter $\bl$ that improves the lower-bound on the data log-likelihood. The lower-bound is tight once $q_{\bl}$ matches $P(\bx|\D)$.

To proceed, we assume a factorized (across individual weights) prior probability parameterized by $\mathbf{m}\equiv\{m_{i\mu}\}$: $P(\bx)=\prod_{i,\mu}\left[\frac{1+m_{i\mu}}{2}\delta_{\xi_{i}^\mu,+1}+\frac{1-m_{i\mu}}{2}\delta_{\xi_{i}^\mu,-1}\right]$ where $\delta_{x,y}$ denotes the Kronecker delta function, and $m_{i\mu}$ defines 
the mean of the synaptic
strength from visible neuron $i$ to hidden one $\mu$ (Fig.~\ref{vrbm})~\cite{Blun-2015,Pbp-2015}. For simplicity, we also parameterize the variational distribution in the same form but with different means specified by $\bl$.
The variational distribution is given by $q_{\bl}(\bx)=\prod_{i,\mu}\left[\frac{1+\lambda_{i\mu}}{2}\delta_{\xi_{i}^\mu,+1}+\frac{1-\lambda_{i\mu}}{2}\delta_{\xi_{i}^\mu,-1}\right]$. Hence the weight uncertainty can be explicitly captured by the variational distribution,
in contrast to the point-estimate in a usual contrast divergence algorithm for continuous weights.
This form was also recently used in supervised learning of perceptron models~\cite{Zecchina-2018,ICLR-2018}.

According to Eq.~(\ref{Pbs}), we can write $\ln P(\D|\bx)$ explicitly and insert it into the lower-bound, then we get
\begin{equation}\label{LB2}
\begin{split}
 {\rm LB}(q_{\bl})&=-{\rm KL}(q_{\bl}(\bx)\|P(\bx))\\
 &+\E\left[\sum_{a,\mu}\ln\cosh(\beta\X)-M\ln Z(\bx)\right].
 \end{split}
\end{equation}
The variational parameter $\bl$ is achieved by optimizing the lower-bound through gradient ascent, defined by $\Delta\bl=\eta\nabla_{\bl}{\rm LB}(q_{\bl})$ where $\eta$ is a learning rate. Before evaluating the gradient,
we need to calculate the regularization and log-likelihood terms. Due to the factorization assumption, the regularization term can be exactly computed as
\begin{equation}\label{KL2}
\begin{split}
 -&{\rm KL}(q_{\bl}(\bx)\|P(\bx))=\sum_{x=\pm1}\sum_{i,\mu}\left[\S(z,y)-\S(z,z)\right],
 \end{split}
\end{equation}
where $z=\frac{1+\lambda_{i\mu}x}{2}$, $y=\frac{1+m_{i\mu}x}{2}$, and $\S(z,y)\equiv z\ln y$.

The expected log-likelihood term is still difficult to estimate without any approximations. However, based on the fact that the variational distribution $q_{\bl}$ is factorized, we assume that
$\X$ involves a sum of a large number of nearly independent random variables and thus follows a Gaussian distribution $\mathcal{N}(G_\mu^a,\Xi_\mu^2)$, where the mean and variance can be computed respectively by
 $G_\mu^a=\frac{1}{\sqrt{N}}\sum_{i\in\partial\mu}\lambda_{i\mu}\sigma_i^a$, and
 $\Xi_\mu^2=\frac{1}{N}\sum_{i\in\partial\mu}(1-\lambda_{i\mu}^2)$, where $i\in\partial\mu$ denotes all incoming neurons into the $\mu$-th hidden neuron. In this form,
the uncertainty of weights on the provided data $\D$ has been incorporated into
this approximation given a large value of $N$.

Given one sensory input, say $\bs^a$, the weighted-sum $\X$ is conditionally independent~\cite{Hinton-2002}.
It then follows that the expected log-likelihood can be approximated by a Monte-Carlo estimation~\cite{MCgrad-2019}:
\begin{equation}\label{Elog}
\begin{split}
 &\E\left[\sum_{a,\mu}\ln\cosh(\beta X_\mu^a)-M\ln Z(\bx)\right]=\\
 &\frac{1}{B_1}\sum_{a,\mu,s}\ln\cosh\left(\beta G_\mu^a+\beta\Xi_\mu z_\mu^s\right)\\
 &-\frac{M}{B_2}\sum_{s}\ln\sum_{\bs}\prod_\mu\cosh\left(\beta G_\mu+\beta\Xi_\mu z_\mu^s\right),
\end{split}
 \end{equation}
where $s$ denotes the index for a Monte-Carlo sample, $z_\mu^s$ is a standard normal random variable, $B_1$ and $B_2$ denote the number of Monte-Carlo samples,
and $G_\mu$ is defined by $G_\mu^a$ without the symbol $a$. Interestingly,
the partition-function part of the expected log-likelihood reduces to an equivalent RBM model (the right panel of Fig.~\ref{vrbm}) whose synaptic connections are now replaced by the variational parameter $\bl$ scaled by $\sqrt{N}$, and
an additional quenched random field is introduced for each individual hidden neuron as $\Xi_\mu z_\mu^s$ due to the fluctuation of the weighted-sum input.
This surprising transformation makes estimation of the original computational hard partition function in Eq.~(\ref{LB2}) tractable, as the cavity method developed for
a RBM~\cite{Huang-2015b} can be directly applied. In brief, a cavity probability or message $P_{i\rightarrow\mu}(\sigma_i$) without considering the $\mu$-th hidden neuron can be 
written into a closed-form iteration~\cite{SM}. The fixed-point messages can be used to estimate the partition function and other thermodynamic quantities related to learning~\cite{SM}.
Another benefit is that the resulting expression in Eq.~(\ref{Elog}) is amenable to the variational inference, i.e., gradient estimation.

One can then derive
the gradient ascent formula to update the variational parameter as $\lambda_{i\mu}^{t+1}=\lambda_{i\mu}^t+\eta\Delta_{i\mu}$, where $t$ denotes the learning step, and $\Delta_{i\mu}$ is decomposed into three parts~\cite{SM}, given by
\begin{equation}\label{grad}
\Delta_{i\mu}=\Delta_{i\mu}^{{\rm Reg}}+\Delta_{i\mu}^{{\rm D}}-\Delta_{i\mu}^{{\rm Eq}},
\end{equation}
where the first term stems from the regularization, namely $\Delta_{i\mu}^{{\rm Reg}}\equiv\sum_{x=\pm1}\frac{x}{2}\left(\ln\frac{1+xm_{i\mu}}{1+x\lambda_{i\mu}}-1\right)$,
the second data-dependent term $\Delta_{i\mu}^{{\rm D}}\equiv\frac{\beta}{B_1\sqrt{N}}\sum_{a,s}\sigma_i^a\tanh(\beta G_\mu^a+\beta\Xi_\mu z_\mu^s)
 -\frac{\beta^2\lambda_{i\mu}}{NB_1}\sum_{a,s}\left[1-\tanh^2\left(\beta G_\mu^a+\beta\Xi_\mu z_\mu^s\right)\right]$, and the final model-dependent term $\Delta_{i\mu}^{{\rm Eq}}$ is estimated from
the equivalent RBM model, given by $\frac{M\beta}{\sqrt{N}B_2}\sum_s\left[C_{i\mu}-\frac{\lambda_{i\mu}z_\mu^s}{\sqrt{N}\Xi_\mu}\hat{m}_\mu\right]$, 
where $C_{i\mu}$ and $\hat{m}_{\mu}$ define the correlation between visible and hidden neurons, and mean activities of hidden neurons, respectively. 
These two thermodynamic quantities are estimated under the Boltzmann measure of the equivalent model $P_{{\rm eq}}(\bs)=\frac{1}{Z_{{\rm eq}}^s}\prod_{\mu}\cosh\left(\beta G_\mu+\beta\Xi_\mu z_\mu^s\right)$ by the message passing procedure.
Note that $|\lambda_{i\mu}|\leqslant1$, and $\xi_i^\mu$ can be decoded as $\xi_i^\mu={\rm sgn}(\lambda_{i\mu})$, so-called maximizer of the posterior marginals~\cite{Nishimori-2001}.

\begin{figure}
     \includegraphics[bb=11 1 586 408,width=0.5\textwidth]{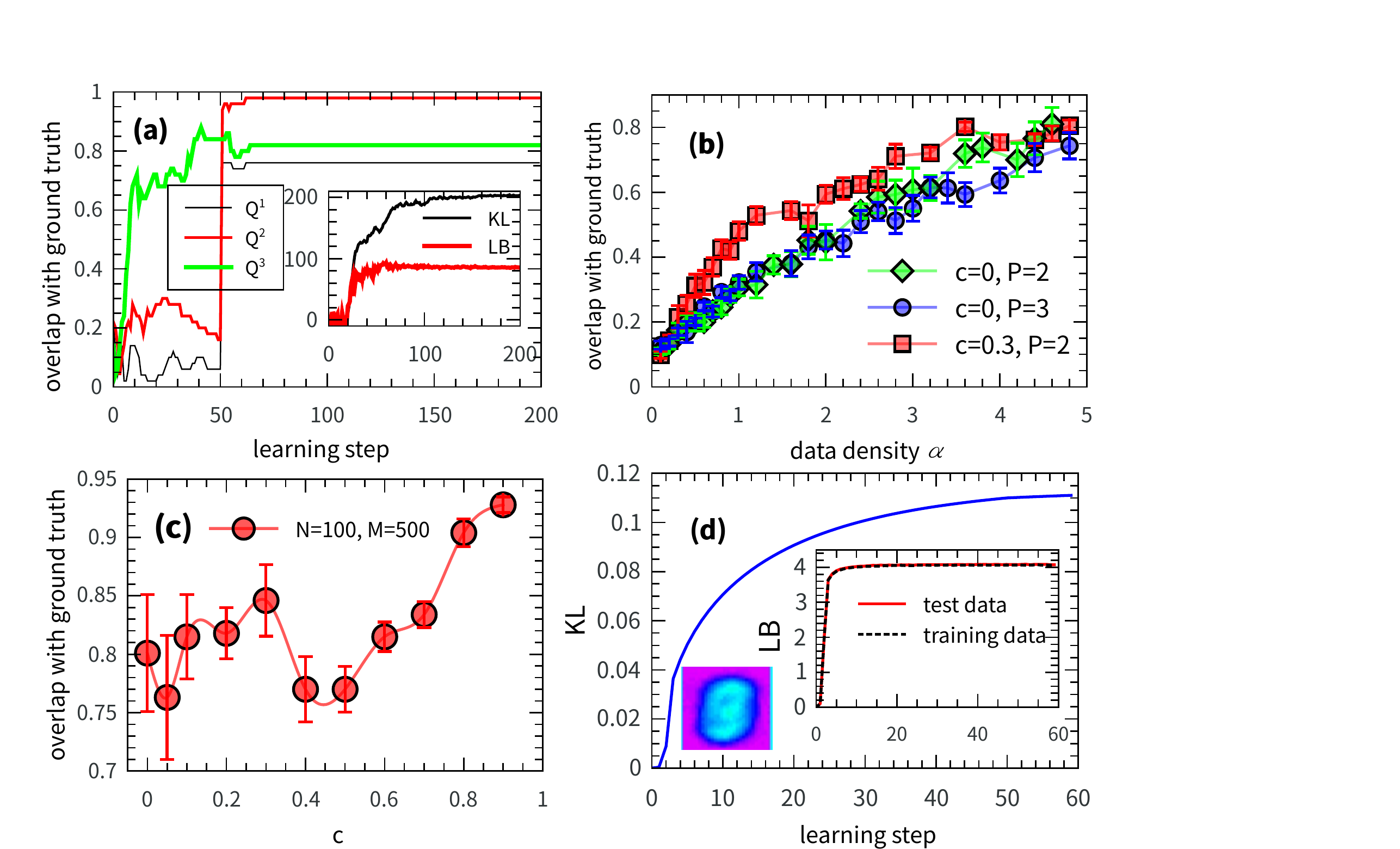}
  \caption{
  (Color online) Learning performance in planted RBM models ($N=100$) and real datasets ($N=784$). The inverse temperature $\beta=1$.
  Each learning step indicates all synapses are updated once. $B_1=B_2=1000$. Here, the Kullback-Leibler divergence refers to ${\rm KL}(q_{\bl}(\bx)\|P(\bx))$, and the approximate lower-bound of the 
  data likelihood (Eqs.~(\ref{LB2}-\ref{Elog})) has been subtracted by its starting value.
  (a) Learning trajectories of RBM models with $P=3$ and orthogonal planted receptive field vectors of three hidden neurons ($M=500$).
  The inset shows evolution of the Kullback-Leibler divergence and approximate lower-bound.
(b) Typical learning performance in planted RBM models as a function of data density $\alpha=\frac{M}{N}$. 
  The learning performance is estimated at the $200$-th step. Each marker in the plot indicates the average over ten independent realizations of the model, 
  with the error bar indicating the standard deviation. The theoretical threshold $\alpha_{{\rm thr}}=1$ for $c=0$~\cite{Huang-2017,Barra-2017} and $\alpha_{{\rm thr}}=0.596$ for $c=0.3$~\cite{Huang-2019}.
  (c) Learning performance versus correlation $c$ with $P=2$ and $M=500$. The average is done over ten independent realizations.
 (d) Learning trajectories (KL and LB per model-parameter) of RBM models with $P=100$ for structured (handwritten digits, thus $N=784$) inputs of $M=1000$ images (another $1000$ images for test).
 $B_1=B_2=500$. A localized variational parameter map (within the receptive field) is also shown.
  }\label{transition}
\end{figure}

\textit{Results.}---
We first test our variational framework on planted RBM models. More precisely, we first generate a RBM model whose synaptic weights are randomly generated with a specified 
correlation level $c$. The correlation-free case of $c=0$ corresponds to the orthogonal weight vectors of hidden neurons. Based on this RBM model with planted ground truth, a collection of data samples is
prepared from Gibbs samplings of the model~\cite{Hinton-2002,Huang-2019}. These data samples are finally used as sensory inputs to our variational learning algorithm, with the goal of reconstructing the ground truth.
The learning performance is measured by the overlap $Q^\mu=\frac{1}{N}\sum_{i=1}^N\xi_i^{\mu,{\rm prd}}\xi_i^{\mu,{\rm plt}}$, quantifying the similarity between predicted and planted weights. 

Typical behavior of the proposed variational mean-field framework is shown in Fig.~\ref{transition}. The variational learning framework is able to recover the ground truth, even in the case of three or more hidden neurons, which was
previously out of reach,
confirming that the variational mean-field theory is capable of capturing the complex interaction among
sensory inputs, synaptic plasticity and neural activities. Most interestingly, there appears a permutation-symmetry-broken phenomenon in inferred 
synaptic weights, which is shown by the observation that when we carry out a permutation operation to receptive fields at an intermediate step (e.g., $50$-th step), the overlap with the ground truth can have a 
significant quantitative change (Fig.~\ref{transition} (a)). The permutation-symmetry-broken phase indeed exists as also theoretically proved in a two-hidden-neuron case~\cite{Huang-2019}. The simulation results of our new variational 
framework add further evidences to this fundamental phenomenon, thereby making testing theoretical predictions of simple models possible in more generic architectures.

Fig.~\ref{transition} (a) shows that the KL divergence between the variational posterior and the prior grows with learning, demonstrating that our variational principle is able to search for a biased 
probability of weights. Note that in the algorithm we do not assume any prior knowledge about the weight vectors, or $m_{i\mu}=0$ for all $(i\mu)$. However, the algorithm itself
takes the variability of the data samples into account,
driving the update of the variational parameter towards the ground truth, as indicated by the saturation of the KL divergence.
In addition, the approximate lower-bound on the data
log-likelihood also grows with learning, suggesting that the variational mean-field framework indeed improves the approximate bound. 

It is clear that in the correlation-free case, the learning threshold does not depend on 
the number of hidden neurons ($P=3$ or $P=2$), in accord with the recent theoretical work~\cite{Huang-2019} which explained that the underlying physics is the partition function factorization.
Furthermore, the correlation level decreases the threshold~\cite{Huang-2019,Hou-2019}, as also verified in current simulation results (Fig.~\ref{transition} (b)). The threshold is rounded by the finite size of the system.
Below the threshold, a random-guess phase ($\bl=0$) dominates the learning. Above the threshold, there appears spontaneous symmetry breaking corresponding to concept-formation in unsupervised learning~\cite{Huang-2016b,Huang-2017}.

Although we restrict the variational distribution to be in the mean-field family, the learning performance shows robustness to different planted correlation levels (Fig.~\ref{transition}(c)),
consistent with popular choices of mean-field posterior in deep learning~\cite{Blun-2015,Pbp-2015,Zecchina-2018,ICLR-2018,Gal-2020}. The variational principle searches for 
the best approximate posterior minimizing the KL divergence in Eq.~(\ref{KL}). It is thus reasonable that by adjusting the variational parameters, given enough data, the true weights can be recovered
such that the probability distance is minimized.

Finally, we apply the proposed framework to the MNIST dataset~\cite{mnist} (Fig.~\ref{transition} (d)). We clearly see that the original uniform prior is not preferred, thus the KL divergence increases until a highly biased (non-uniform)
probability of weight configurations is reached. The learned highly-structured receptive fields can provide informative priors for fine-tuning deep neural networks~\cite{Hinton-2006b,ICLR-2018}.
A detailed study is left for our future works.
Meanwhile, the approximate lower-bound for both training and test (unseen) datasets increases until saturation. 
Therefore, our framework is also promising in studying structured real dataset, especially for extracting or verifying principles of unsupervised learning~\cite{Huang-2016b,Huang-2017,Barra-2017,Monasson-2017,Huang-2018,Barra-2018,Dece-2019,Huang-2019,Hou-2019}.

\textit{Conclusion.}---
Although training and understanding of neural networks with low-precision weights and activations becomes increasingly important~\cite{Hubara-2017,Zecchina-2018},
a unified theoretical framework to treat sensory inputs, synapses and neural activity together is still lacking, and thus the conceptual advance contributed by
our variational mean-field theory provides a route towards an in-depth understanding of unsupervised learning in a principled probabilistic framework. In this framework,
the synaptic weight is no longer treated to be deterministic, but rather, it is subject to a variational distribution where the variational parameter, mean synaptic activity level, is adjusted by
an expectation-maximization-like mechanism~\cite{rep-2019}. The gradient ascent of the approximate lower-bound behaves as an M-step at the synaptic activity level, while the E-step at the neural activity level is carried out by a fully-distributed message passing procedure on an equivalent neural network whose network parameters are determined in turn
by the variational parameters. 
These two steps as a key to unsupervised learning are unified into a single equation (Eq.~(\ref{grad})), providing insights on how data (or sensory inputs), synapses and neurons shape the unsupervised learning. Interestingly, this variational principle marrying gradient ascent and message passing shares
the similar spirit to the free-energy principle proposed for explaining action, perception and learning in the brain~\cite{Karl-2010}.

The proposed variational mean-field theory for challenging discrete synapses in unsupervised learning encourages developing theory-grounded neuromorphic 
algorithms on neural networks with low-precision yet robust synapses and activations~\cite{morphic-2015,Hubara-2017}. Last but not the least, this framework opens a new way to test the hypothesis that unsupervised learning can be
interpreted as breaking different types of inherent symmetry in a data-driven spontaneous manner~\cite{Huang-2019,Hou-2019}, in a general setting, i.e., neural network architectures with arbitrarily many hidden neurons, and hidden layers.

\section*{Acknowledgments}

This research was supported by the start-up budget 74130-18831109 of the 100-talent-
program of Sun Yat-sen University, and the NSFC (Grant No. 11805284).

\setcounter{figure}{0}    
\renewcommand{\thefigure}{S\arabic{figure}}
\renewcommand\theequation{S\arabic{equation}}
\setcounter{equation}{0}  

\onecolumngrid
\appendix

\section*{Supplemental Material}

\section{Mean-field estimation of the partition function of the equivalent model for learning}
\label{SM-a}
 In this supplemental material, we briefly summarize the mean-field estimation of the partition function of the equivalent model as follows. Technical details to derive these results based on the cavity method have been given in a series of 
works~\cite{Huang-2015b,Huang-2017}. This estimation provides a practical procedure called message passing working on single instances of the RBM, which is efficient for a practical learning.
We first write the partition function for a Monte-Carlo sample indexed by $s$ as follows,
\begin{equation}\label{Zs}
 Z_{{\rm eq}}^s=\sum_{\bs}\prod_\mu\cosh\left(\beta G_\mu+\beta\Xi_\mu z_\mu^s\right),
\end{equation}
where $G_\mu$ and $\Xi_\mu$ are defined by $G_\mu=\frac{1}{\sqrt{N}}\sum_{i\in\partial\mu}\lambda_{i\mu}\sigma_i$ and
 $\Xi^2_\mu=\frac{1}{N}\sum_{i\in\partial\mu}(1-\lambda_{i\mu}^2)$, respectively. Therefore, in the equivalent RBM, $\frac{\lambda_{i\mu}}{\sqrt{N}}$ acts as a continuous weight,
 and $H_\mu\equiv\Xi_\mu z^s_\mu$ acts a quenched random field applied to the $\mu$-th hidden neuron in the equivalent RBM.

First, the interaction between visible and hidden neurons is of the order $\mathcal{O}(\frac{1}{\sqrt{N}})$ and thus weak, then the cavity approximation taking only correlations around a factor in the product of Eq.~(\ref{Zs})
leads to the following self-consistent cavity iteration:
\begin{subequations}\label{bp1}
 \begin{align}
 P_{i\rightarrow\nu}(\sigma_i)&=\frac{1}{Z_{i\to\nu}}\prod_{\mu\in\partial i\backslash\nu}\hat{P}_{\mu\to i}\label{bp1a} \\ 
 \hat{P}_{\mu\rightarrow i}(\sigma_i)&=\sum_{\{\sigma_j|j\in\partial\mu\backslash i\}}\cosh\Biggl(\sum_{j\in\partial\mu\backslash i}\frac{\beta\lambda_{j\mu}\sigma_j}{\sqrt{N}}
 +\frac{\beta\lambda_{i\mu}\sigma_i}{\sqrt{N}}+\beta H_\mu\Biggr)\prod_{j\in\partial\mu\backslash i}P_{j\to\mu}(\sigma_j),\label{bp1b}
  \end{align}
\end{subequations}
where $P_{i\to\nu}(\sigma_i)$ denotes the cavity probability for $\sigma_i$ given that only the $\nu$-th hidden neuron is removed (this is what the cavity means), while
$\hat{P}_{\mu\to i}(\sigma_i)$ summarizes all contributions around the factor $\mu$ when the value of $\sigma_i$ is given. $Z_{i\to\nu}$ is a normalization constant for $P_{i\to\nu} (\sigma_i)$.

Note that the concept of cavity is required for using the factorization of the joint probability of neural activities to
write down the above closed-form iterative equation on the graphical model where the visible neuron acts as a variable node, and the hidden neuron acts as a factor/function node in a factor-graph representation of 
the model~\cite{Huang-2015b,Frey-2001}. The resulting iterative equation is also named the belief propagation in computer science community~\cite{Yedidia-2005}.
However, it is still not easy to evaluate $\hat{P}_{\mu\to i} (\sigma_i)$ without further approximations, due to the intractable summation in Eq.~(\ref{bp1b}).
By noting that the term in Eq.~(\ref{bp1b})---$\sum_{j\in\partial\mu\backslash i}\frac{\beta\lambda_{j\mu}\sigma_j}{\sqrt{N}}$ involves a sum of a large number of nearly-independent random terms,
one can apply the central-limit-theorem as an approximation whose justification can be made in practical learning experiments. Then the original intractable summation can be replaced by an integral involving Gaussian random variables.
Therefore, the cavity magnetization $m_{i\to\nu}=\sum_{\sigma_i=\pm1}\sigma_iP_{i\to\nu}(\sigma_i)$ under the measure of
the cavity probability can be readily obtained as
\begin{subequations}\label{bp2}
 \begin{align}
 m_{i\rightarrow\nu}&=\tanh\left(\sum_{\mu\in\partial i\backslash\nu}u_{\mu\rightarrow i}\right),\\
 u_{\mu\rightarrow i}&=\tanh^{-1}\left(\tanh(\beta\cmui+\beta H_\mu)\tanh(\beta\lambda_{i\mu}/\sqrt{N})\right),
  \end{align}
\end{subequations}
where $\mu\in\partial i\backslash\nu$ indicates all factors around the visible neuron $i$ excluding the factor indexed by $\nu$. $m_{i\rightarrow\mu}$ can be interpreted as the message passing from visible neuron to hidden neuron (Fig.~1), while
$u_{\mu\rightarrow i}$ denotes the cavity bias interpreted as the message passing from hidden neuron to visible neuron. The mean of the Gaussian distribution approximation under the central-limit-theorem is given by
$\cmui\equiv\frac{1}{\sqrt{N}}\sum_{j\in\partial\mu\backslash i}\lambda_{j\mu}m_{j\rightarrow\mu}$ collecting all messages except those from
$i$ around the factor
$\mu$, weighted by their corresponding variational parameters, and $H_\mu$ denotes the quenched random field expressed as $\Xi_\mu z_\mu^s$.

Because of weak interactions, this message passing equation is able to converge even in a few steps. Then
the log-partition-function (so-called free energy) can be readily constructed based on the visible neuron's contribution $F_i$ and the factor's contribution $F_\mu$, as $\ln Z=\sum_i F_i-(N-1)\sum_\mu F_\mu$ where the subtraction removes the double counting of the contribution from
the first term~\cite{cavity-2001}. $F_i$ and $F_\mu$ are given respectively by
\begin{subequations}\label{free}
 \begin{align}
 F_i&=\ln\sum_{\sigma_i}\prod_{\mu\in\partial i}\hat{P}_{\mu\to i}(\sigma_i)=\sum_{\mu\in\partial i}\left[\beta^2\Lmui^2/2+\ln\cosh\left(\beta\cmui+\beta H_\mu+\beta\lambda_{i\mu}/\sqrt{N}\right)\right]+\ln\left(1+\prod_{\mu\in\partial i}e^{-2u_{\mu\rightarrow i}}\right),\\
 F_{\mu}&=\ln\int d\varrho_{\mu}\cosh\left(\beta\varrho_\mu+\beta H_{\mu}\right)\mathcal{N}(\varrho_\mu;\cmu,\Lmu^2)=\beta^2\Lmu^2/2+\ln\cosh\left(\beta\cmu+\beta H_\mu\right),
  \end{align}
\end{subequations}
where $\Lmui^2\equiv\frac{1}{N}\sum_{j\in\partial\mu\backslash i}\lambda_{j\mu}^2(1-m_{j\rightarrow\mu}^2)$, $\Lmu^2\equiv\frac{1}{N}\sum_{j\in\partial\mu}\lambda_{j\mu}^2(1-m_{j\rightarrow\mu}^2)$, 
and $\cmu\equiv\frac{1}{\sqrt{N}}\sum_{i\in\partial\mu}\lambda_{i\mu}m_{i\rightarrow\mu}$. Note that $\cmu$ and $G_\mu$ are intrinsically different, because the former describes the equilibrium properties of
the equivalent RBM with fixed $\lambda_{i\mu}$ (the right panel of Fig.~1 in the main text), while the latter captures the statistics under the variational distribution given the sensory input (the left panel of Fig.~1 in the main text).

We finally remark that the free energy is computed using the cavity approximation, the lower-bound of the data log-likelihood is thus an approximate estimation of 
the original intractable lower-bound (Eq.~(5) in the main text).


\section{Derivations of gradients of the data log-likelihood lower-bound}
\label{SM-b}
Finally, let us evaluate the gradient of the lower-bound. First, the gradient of the regularization term with respect to the variational parameter is obtained as
\begin{equation}\label{grad1}
 -\frac{\partial}{\partial\lambda_{i\mu}}{\rm KL}(q_{\bl}(\bx)\|P(\bx))=\sum_{x=\pm1}\frac{x}{2}\left(\ln\frac{1+xm_{i\mu}}{1+x\lambda_{i\mu}}-1\right).
\end{equation}
It is clear that this term vanishes when the variational distribution is exactly matched to the prior.
Second, the gradient of the first term in the expected log-likelihood (Eq.~(7) in the main text) can be written as
\begin{equation}\label{grad2}
\begin{split}
 \frac{\partial}{\partial\lambda_{i\mu}}\E\left[\sum_{a,\mu}\ln\cosh(\beta\X)\right]&\simeq\frac{\beta}{B_1\sqrt{N}}\sum_{a,s}\sigma_i^a\tanh(\beta G_\mu^a+\beta\Xi_\mu z_\mu^s)\\
 &-\frac{\beta^2\lambda_{i\mu}}{NB_1}\sum_{a,s}\left[1-\tanh^2\left(\beta G_\mu^a+\beta\Xi_\mu z_\mu^s\right)\right].
 \end{split}
\end{equation}
Lastly, the gradient of the second term in the expected log-likelihood can be derived as
\begin{equation}\label{grad3}
\begin{split}
 \frac{\partial}{\partial\lambda_{i\mu}}\E\ln Z(\bx)&\simeq\frac{\beta}{\sqrt{N}B_2}\sum_s\left<\sigma_i\tanh(\beta G_\mu+\beta\Xi_\mu z_\mu^s)\right>\\
 &-\frac{\beta\lambda_{i\mu}}{NB_2}\sum_{s}\left[\frac{z_\mu^s}{\Xi_\mu}\left<\tanh\left(\beta G_\mu+\beta\Xi_\mu z_\mu^s\right)\right>\right],\\
 &=\frac{\beta}{\sqrt{N}B_2}\sum_s\left[C_{i\mu}-\frac{\lambda_{i\mu}z_\mu^s}{\sqrt{N}\Xi_\mu}\hat{m}_\mu\right],
 \end{split}
\end{equation}
where the expectation $\left<\cdot\right>$ means an average under the Boltzmann measure of the equivalent model $P_{{\rm eq}}(\bs)=\frac{1}{Z_{{\rm eq}}^s}\prod_{\mu}\cosh\left(\beta G_\mu+\beta\Xi_\mu z_\mu^s\right)$,
$C_{i\mu}$ and $\hat{m}_{\mu}$ thus define the correlation between visible and hidden neurons, and mean activities of hidden neurons, respectively. Interestingly, these two macroscopic thermodynamic quantities can be evaluated 
from the fixed point of the message passing equation (Eq.~(\ref{bp2})). Interested readers can find technical details in our previous work~\cite{Huang-2015b}. Here we summarize the result as follows,
\begin{subequations}\label{corre}
 \begin{align}
 m_i&=\tanh\left(\sum_{\mu\in\partial i}u_{\mu\rightarrow i}\right),\\
 \hat{m}_\mu&=\int Dz\tanh(\beta\tcmu+\beta H_\mu+\beta\tLmu z),\\
 C_{i\mu}&=\hat{m}_\mu m_i+\frac{\beta\lambda_{i\mu}}{\sqrt{N}}(1-m_i^2)A_\mu,\\
 A_\mu&=1-\int Dz\tanh^2(\beta\tcmu+\beta H_\mu+\beta\tLmu z),
  \end{align}
\end{subequations}
where $Dz\equiv e^{-z^2/2}/\sqrt{2\pi}dz$, $\tcmu\equiv\frac{1}{\sqrt{N}}\sum_{i\in\partial\mu}\lambda_{i\mu}m_i$, and $\tLmu^2\equiv\frac{1}{N}\sum_{i\in\partial\mu}\lambda_{i\mu}^2(1-m_i^2)$.
$m_i$ is the magnetization (mean activity) of the visible neuron and can thus be read off from the fixed point of the message passing equation.
$\hat{m}_\mu\equiv\left<\tanh\Biggl(\sum_{i\in\partial\mu}\frac{\beta\lambda_{i\mu}\sigma_i}{\sqrt{N}}+\beta H_\mu\Biggr)\right>$ which can be easily estimated by using
the central-limit theorem once again and then calculating a Gaussian integral.

The computation of the correlation $C_{i\mu}$ is a bit tricky. We first define an auxiliary quantity $\hat{C}_{i\mu}=C_{i\mu}-\hat{m}_\mu m_i$, then compute
the summation $\sum_{i\in\partial\mu}\frac{\beta\lambda_{i\mu}}{\sqrt{N}}\hat{C}_{i\mu}$ as follows,
\begin{equation}\label{corre2}
\begin{split}
 \sum_{i\in\partial\mu}\frac{\beta\lambda_{i\mu}}{\sqrt{N}}\hat{C}_{i\mu}&=\left<\tanh(\beta\varrho_\mu+\beta H_\mu)\beta\varrho_\mu\right>-\hat{m}_\mu\beta\tcmu,\\
 &=\int d\varrho_\mu\mathcal{N}(\varrho_\mu;\tcmu,\tLmu^2)\tanh(\beta\varrho_\mu+\beta H_\mu)(\beta\varrho_\mu)-\beta\hat{m}_\mu\tcmu,\\
 &=\int Dz\tanh(\beta\tcmu+\beta H_\mu+\beta\tLmu z)(\beta\tcmu+\beta\tLmu z)-\beta\hat{m}_\mu\tcmu,\\
 &=\beta^2\tLmu^2\int Dz\Biggl(1-\tanh^2\Bigl(\beta\tcmu+\beta H_\mu+\beta\tLmu z\Bigr)\Biggr),
 \end{split}
\end{equation}
where we define $\varrho_\mu\equiv\sum_{i\in\partial\mu}\frac{\lambda_{i\mu}}{\sqrt{N}}\sigma_i$, and $C_{i\mu}\equiv\left<\tanh(\beta\varrho_\mu+\beta H_\mu)\sigma_i\right>$ by definition.
From the last equality of Eq.~(\ref{corre2}), we can easily get 
$\hat{C}_{i\mu}=\frac{\beta\lambda_{i\mu}}{\sqrt{N}}(1-m_i^2)A_\mu$, from which the correlation $C_{i\mu}$ is obtained.

Collecting all three parts of the gradient, we arrive at the gradient ascent formula to update the variational parameter as $\lambda_{i\mu}^{t+1}=\lambda_{i\mu}^t+\eta\Delta_{i\mu}$, where $t$ denotes the learning step, and $\Delta_{i\mu}$ is given by
\begin{equation}\label{gradsm}
\begin{split}
\Delta_{i\mu}&=\sum_{x=\pm1}\frac{x}{2}\left(\ln\frac{1+xm_{i\mu}}{1+x\lambda_{i\mu}}-1\right)+\frac{\beta}{B_1\sqrt{N}}\sum_{a,s}\sigma_i^a\tanh(\beta G_\mu^a+\beta\Xi_\mu z_\mu^s)\\
 &-\frac{\beta^2\lambda_{i\mu}}{NB_1}\sum_{a,s}\left[1-\tanh^2\left(\beta G_\mu^a+\beta\Xi_\mu z_\mu^s\right)\right]-\frac{M\beta}{\sqrt{N}B_2}\sum_s\left[C_{i\mu}-\frac{\lambda_{i\mu}z_\mu^s}{\sqrt{N}\Xi_\mu}\hat{m}_\mu\right].
 \end{split}
\end{equation}
Note that $|\lambda_{i\mu}|\leqslant1$, and $\xi_i^\mu$ can be decoded as $\xi_i^\mu={\rm sgn}(\lambda_{i\mu})$.

We finally remark that although the message passing algorithm converges fast to evaluate the equilibrium properties of the dual RBM, there exist other methods, e.g., Gibbs sampling~\cite{Hinton-2002} or
high-temperature expansion~\cite{Krzakala-2018}, for achieving the same goal.



\end{document}